\pdfoutput=1

\documentclass[11pt]{article}

\usepackage[final]{styles/acl}

\usepackage{times}
\usepackage{latexsym}

\usepackage[T1]{fontenc}

\usepackage[utf8]{inputenc}

\usepackage{microtype}

\usepackage{inconsolata}

\usepackage{graphicx}
\usepackage{tabularx}
\usepackage{listings}
\usepackage{amssymb}
\usepackage{soul}

\title{NeMo-Inspector: A Visualization Tool for LLM Generation Analysis}

\author{Daria Gitman \and Igor Gitman \and Evelina Bakhturina \\
        NVIDIA Corporation,\\United States\\
\{dgitman, igitman, ebakhturina\}@nvidia.com}

\begin{document}
\maketitle
\begin{abstract}
Adapting Large Language Models (LLMs) to novel tasks and enhancing their overall capabilities often requires large, high-quality training datasets. Synthetic data, generated at scale, serves a valuable alternative when real-world data is scarce or difficult to obtain. However, ensuring the quality of synthetic datasets is challenging, as developers must manually inspect and refine numerous samples to identify errors and areas for improvement. This process is time-consuming and requires specialized tools. We introduce NeMo-Inspector, an open-source tool designed to simplify the analysis of synthetic datasets with integrated inference capabilities. We demonstrate its effectiveness through two real-world cases. Analysis and cleaning of the synthetically generated GSM-Plus dataset with NeMo-Inspector led to a significant decrease in low-quality samples from 46.99\% to 19.51\%. The tool also helped identify and correct generation errors in OpenMath models, improving accuracy by 1.92\% on the MATH dataset and by 4.17\% on the GSM8K dataset for a Meta-Llama-3-8B model fine-tuned on synthetic data generated from Nemotron-4-340B. 
\end{abstract}

\begin{table*}
  \centering
  \begin{tabular}{lccccc}
  \hline
\textbf{Features} & \textbf{NeMo} & \textbf{Lilac} & \textbf{LLM-} & \textbf{LLM-} & \textbf{KNIME} \\
  & \textbf{Inspector} & & \textbf{Attributor}& \textbf{Comparator} &  \\
\hline
Syntax formatting & \checkmark & Markdown only & X & X & X \\
Streamlined Inference & \checkmark & X & X & X & X  \\
Multiple generations & \checkmark & X & X & Comparing only & X \\
Custom functions & \checkmark & X & X & \checkmark & \checkmark \\
\hline
\end{tabular}
\caption{Comparison of the NeMo-Inspector with existing freely available tools for data analysis. ``Streamlined Inference'' refers to the ability to perform inference on models without requiring scripting or model weight uploads. ``Multiple generations'' refers to the prepared workflow for simultaneous analysis of both ``homogeneous'' and ``heterogeneous'' generations.}
\label{table:all_tools_comparison}
\end{table*}

\section{Introduction}
Large Language Models (LLMs) have demonstrated remarkable capabilities across various applications. Recently, instead of limited and difficult-to-obtain real-world data, synthetic datasets have become widely popular. However, the quality of the synthetic data cannot be guaranteed. Developers spend a lot of time analyzing model predictions and tuning the prompt to find areas for potential improvements.

This labor-intensive process underscores the need for a tool that can streamline dataset examination, simplify prompt adjustment, and ultimately expedite the refinement of model behaviors.
There are several tools available that offer valuable features for dataset analysis, but each has its own limitations. 
KNIME~\citep{knime} includes features for comprehensive data analysis, but lacks the convenience of sample-by-sample inspection and text formatting. Lilac~\citep{lilac} focuses on the analysis of LLM datasets but does not support inference, LaTeX formatting, or multi-generation comparisons. LLM-Attributor~\citep{llm_atributor} is a specialized tool for understanding the influence of training data on the generated text; however, it is not suitable for the analysis of independent generations. Finally, LLM-Comparator~\citep{llm_comparator} allows side-by-side comparisons of model outputs but lacks integrated inference capabilities, manual editing, and handling for homogeneous multi-generation scenarios, allowing not only side-by-side but also cross-generation analysis.
Table~\ref{table:all_tools_comparison} shows the comparison of NeMo-Inspector with other tools for data analysis.

To fill this research gap, we introduce NeMo-Inspector~\footnote{\url{https://github.com/NVIDIA/NeMo-Inspector}} - an open-source tool released under an Apache 2.0 License. NeMo-Inspector provides the following features:
\begin{enumerate}
    \item \textbf{Support for Different Types of Generations}. 
    NeMo-Inspector facilitates systematic analysis of LLM outputs by categorizing them into \textit{homogeneous} and \textit{heterogeneous} generations. \textit{Homogeneous} generations, which are produced by varying random seeds within the same model and parameter settings, can be grouped for collective analysis. This approach enables the computation of aggregate statistics, such as the proportion of correct responses. In contrast, \textit{heterogeneous} generations, originating from distinct models, parameter configurations, or tasks, are inherently disparate and not directly comparable as a unified set. These outputs are examined individually or side-by-side to elucidate differences in performance and behavior.
    \item \textbf{Custom functions}. The NeMo-Inspector supports user-defined Python functions to enable customized filtering, sorting, editing, and statistical analysis.

    \item \textbf{Manual editing}. Reviewing synthetic data typically requires examining a large number of individual samples. Incorporating notes or labels during this process enhances contextual understanding and enables a more detailed and comprehensive analysis.

    \item \textbf{Streamlined inference through entry points for Different Formats}. The tool enables interactive inference, facilitating rapid prompt engineering. This functionality is powered by NeMo-Skills~\footnote{\url{https://github.com/Kipok/NeMo-Skills}} and supports the following formats: NeMo~\cite{nemo}, TensorRT-LLM~\footnote{\url{https://github.com/NVIDIA/TensorRT-LLM}}, vLLM~\cite{kwon2023efficient}, and inference servers that implement the OpenAI API~\footnote{\url{https://platform.openai.com/docs/overview}}. 
    \item \textbf{Flexible visualization}. Poorly formatted content, particularly large text blocks, can hinder comprehension, especially when dealing with specialized datasets, such as those containing complex mathematical formulas, code snippets or domain-specific data. The NeMo-Inspector supports Markdown, LaTeX formatting and Python syntax highlighting. 
    
\end{enumerate}

The rest of the paper is organized as follows.
Section~\ref{section:features} provides an overview of the key features of the tool. Section~\ref{section:applications} highlights common use cases of the NeMo-Inspector. Section~\ref{section:results} presents the results from practical applications of the tool. Finally, sections~\ref{section:conclusion} and~\ref{section:limitations} discuss the overall conclusion and tool's limitations.

\section{Tool overview}
\label{section:features}
The tool includes two pages that can be used separately or together: \textit{the Inference page}, which focuses on interactive prompt design, and \textit{the Analyze page}, which provides a comprehensive examination of the model's outputs.

By integrating both inference and LLM generation analysis within a single tool, users can efficiently test hypotheses and gain insights from generation analysis in real time.

\begin{figure*}[t]
  \includegraphics[width=\linewidth]{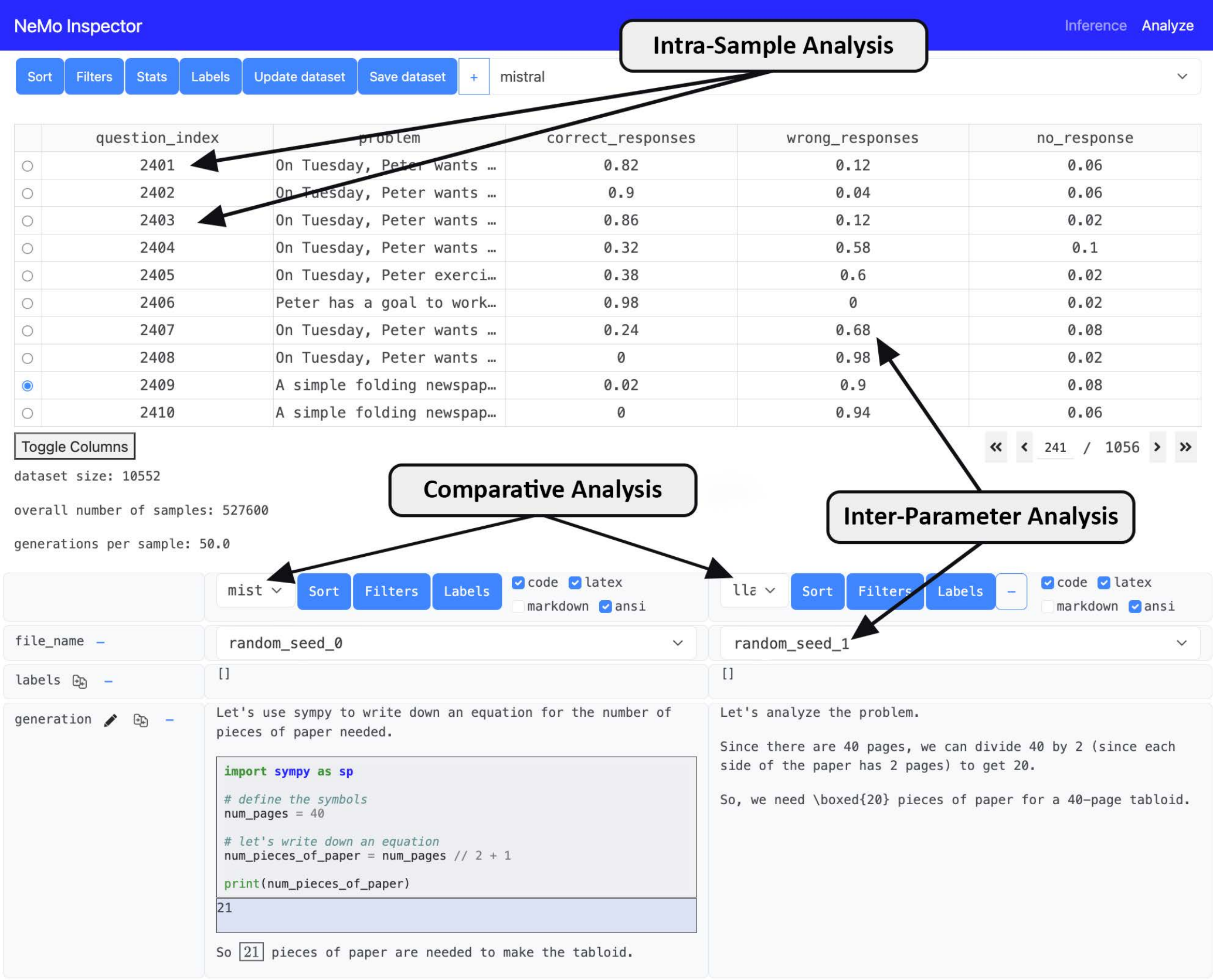}
  \caption{Comprehensive data analysis with the Analyze page of NeMo-Inspector. The tool offers a holistic data view through three complementary perspectives: Intra-Sample Analysis, Comparative Analysis, and Inter-Parameter Analysis.}
  \label{fig:3d_analysis}
\end{figure*}

\subsection{Inference page}
The inference page offers two modes: ``Prompt-based'' and ``Templates-based'', allowing users to set inference parameters and interact with the model. The distinction between the modes lies in the prompt structure.

In ``Prompt-based'' mode, the user is responsible for writing the entire prompt, including all special tokens, query and few shot examples. In contrast, the ``Templates-based'' mode leverages predefined templates, allowing users to select or customize a template by adding placeholders that will later be automatically filled in.

For example, a template for few-shot examples can be structured as:
\begin{verbatim}
Problem: {question}; Solution: {solution}
\end{verbatim}
Here, \texttt{\{question\}} and \texttt{\{solution\}} serve as placeholders that will be automatically replaced with specific problem-solution pairs.

The ``Prompt-based'' mode is designed for exploring a single question, while the ``Templates-based'' mode allows us to experiment with different questions while keeping the settings fixed. Users can set up the questions manually or extract them from the chosen JSON Lines file.

\subsection{Analyze page}
The Analyze page provides features to effectively examine LLM generations and supports any text dataset presented in JSON Lines format. 

NeMo-Inspector provides features to examine many individual samples efficiently. Each sample is presented in a reader-friendly format, i.e. with Markdown and LaTeX rendering, and Python syntax highlighting.
Additionally, the tool enables manual editing, labeling, and highlighting of differences within the selected samples.

The tool offers simultaneous analysis from three perspectives (Figure ~\ref{fig:3d_analysis}):
\begin{itemize}
\item Intra-Sample Analysis. This approach involves evaluating each sample independently, focusing on its individual characteristics rather than comparing it to other samples. For a given model and task, this analysis enables a detailed examination of the sample's input, the corresponding model output, and any associated meta-information on a case-by-case basis.

\item Comparative Analysis. The tool enables side-by-side comparisons of LLM predictions under different inference configurations or models. For example, users can compare model accuracies based on Chain-of-Thought~\cite{cot} prompting versus least-to-most prompting~\cite{least_to_most}.
\item Inter-Parameter Analysis. This involves analysis of model predictions across multiple inference runs, including aggregate statistics, such as calculating metrics based on majority voting or other summary measures.

\end{itemize}

Users can write Python functions to calculate statistics, sort and filter samples across different dimensions of the data, enabling tailored and detailed analysis.

\section{Usage scenarios}
Most of the following applications are based on mathematical datasets, as these were the primary focus of our work. However, the tool is not restricted to mathematical benchmarks and can be effectively applied to any other text datasets. The only requirement is that the datasets be in JSON Lines format.
\label{section:applications}
\subsection{Synthetic data quality check}
\label{gsm_plus}
\subsubsection{Setup}
In this section, we demonstrate the use of the NeMo-Inspector tool for performing data quality checks on the synthetically generated GSM-Plus~\citep{gsm-plus} dataset. GSM-Plus is an adversarially modified version of the GSM8K~\citep{gsm8k} dataset, which consists of human-written grade school math word problems. The GSM-Plus dataset was specifically designed to assess the robustness of LLMs under eight different problem perturbations.

In this example, we work with the following two perturbations (Table~\ref{table:perturbations}):
\begin{itemize}
    \item  \textit{``adding operation''} - modifies the seed question by incorporating additional statements. 
    \item  \textit{``numerical substitution''} - replaces numerical values with another number that has the same number of digits, e.g., replacing $30$ with $80$.
\end{itemize}

\begin{table*}[t]
  \centering
  \begin{tabularx}{\textwidth}{lXc}
    \hline
    \textbf{Perturbation Type} & \textbf{Question} & \textbf{Answer} \\
    \hline
Original & 
    Ali had \$21. Leila gave him half of her \$100. How much does Ali have now? & 71 \\
    \hline 
    Adding Operation & Ali had \$21. Leila gave him half of her \$100. \textbf{After that, Ali spent \$15 on a book and then his friend, John, gave him a quarter of his \$80}. How much does Ali have now?
    & 76 \\
    \hline
    Numerical Substitution & Ali had \textbf{\$30}. Leila gave him half of her \textbf{\$120}. How much does Ali have now? & 90 \\
    \hline
  \end{tabularx}
  \caption{An example of a GSM8K question and its corresponding perturbed versions from the GSM-Plus dataset.}
  \label{table:perturbations}
\end{table*}

 For the analysis, we use the Meta-Llama-3-70B-Instruct (Llama-3-70B) model~\citep{llama}. We sample 50 solutions for each problem with different random seeds, \textit{temperature} \(= 0.7\), \textit{top\_p} \(= 0.95\), and eight few shot examples~\citep{cot} for both GSM8K and GSM-Plus datasets. A larger number of generations enables a more thorough analysis to understand the stability of the model's predictions.

 \subsubsection{Statistics}
 
The tool's Inter-Parameter Analysis enables the aggregation of metrics across different generations (see Figure~\ref{fig:3d_analysis}). In this example, the aggregation is applied across multiple random seeds.

The tool currently includes predefined inter-parameter statistics, such as the ratio of correct, incorrect, or empty responses. Additionally, users have the ability to extend these statistics with custom metrics of their own design.

To identify potential low-quality questions in the GSM-Plus dataset, we introduce a statistic - ``persistence'' (Listing~\ref{lst:custom_stats} ) - indicating the model's confidence in its answers. 

\begin{figure*}[t]
\begin{lstlisting}[language=Python, caption=An example of a custom statistic that counts the maximum number of samples with an identical answer., captionpos=b, frame=single,label=lst:custom_stats]
from collections import Counter
def persistence(datas):
    counter = Counter([data["predicted_answer"] for data in datas])
    return counter.most_common(1)[0][1]
{"persistence": persistence}
\end{lstlisting}
\end{figure*}
We define ``persistence'' 
as the maximum number of generations in which the model produces identical answers to a given question. For instance, if the model generates the answer $A_1$ in $n_1$ run, answer $A_2$ in $n_2$ run, and answer $A_m$ in $n_m$ run, the ``persistence'' is equal to $max(n_1, n_2, ..., n_m)$. %

Questions that consistently yield the same incorrect responses may indicate potential issues with the quality of GSM-Plus samples.

\subsubsection{Sorting} 
Next, we utilize the aggregated statistics to identify and analyze the most challenging questions for the model. The data is sorted to prioritize questions with the lowest accuracy rates and highest persistence levels. Upon examining the top 10 questions, we observe that some samples contain two question marks, which leads to the model's uncertainty regarding which one to address.

\subsubsection{Filtering, labeling and batch editing}
We can identify all data points containing two question marks, and then consider several options for handling them. One approach is to apply a batch editing and automatically remove the extra questions to fix the samples. However, this method carries the risk that the expected answer may no longer be correct, as it could be tied to the removed question. Alternatively, we can label or filter out all problems with two question marks, marking them as ``bad quality'', and save the dataset accordingly.

\subsubsection{Comparison}
\begin{figure*}[t]
  \includegraphics[width=\linewidth]{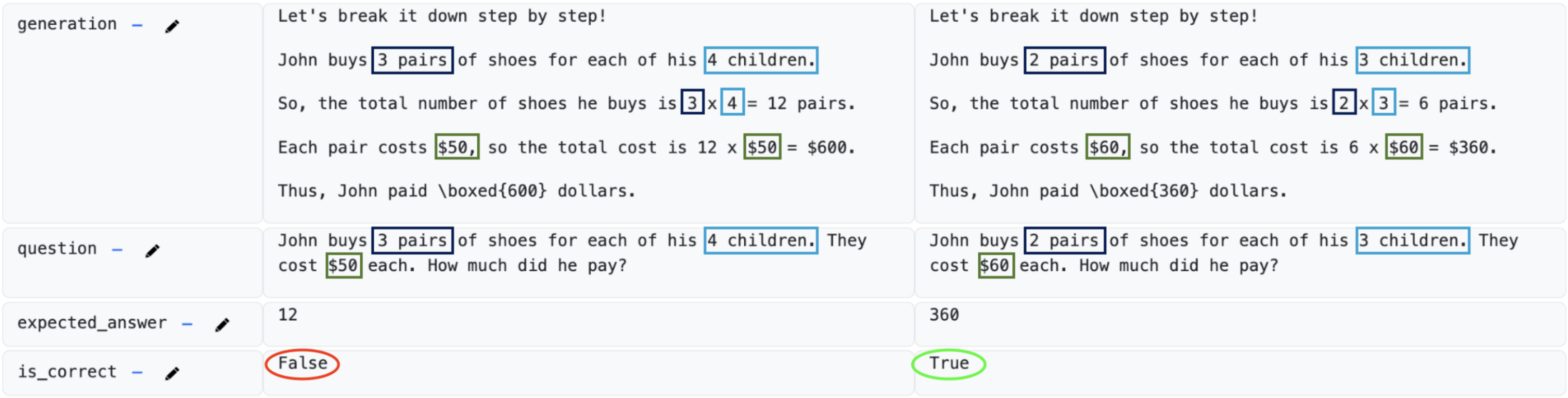}
  \caption{An example of similar solutions with varying correctness illustrates the incorrect expected answer. On the left is a question from the GSM-Plus dataset, and on the right is a question from the GSM8K dataset.}
  \label{fig:comparing}
\end{figure*}

The Comparative Analysis functionality of the tool allows developers to compare model predictions side by side. For example, when analyzing the ``numerical substitution'' category, we compare LLM predictions on the GSM-Plus dataset to those on the corresponding unperturbed GSM8K samples. We expect the reasoning to remain the same as this perturbation type affects only numbers. However, we detect samples with two identical solutions for the perturbed question and the original, but only one of the solutions leads to the expected answer. This suggests that these samples have an incorrect expected answer field (Figure~\ref{fig:comparing}). We can also label or exclude such samples with the NeMo-Inspector tool.

\subsection{Annotation of LLM generations}
\subsubsection{Labeling}
The NeMo-Inspector supports Intra-Sample analysis, allowing for the examination of individual generations. An example of this analysis is data annotation.

To reveal the common problems of the model, we can manually go through a small number of samples and label them according to some predefined taxonomy. 

For example, consider a dataset of chatbot customer service responses generated by an LLM. Researchers manually label each response to highlight instances that are too vague, fail to address the customer's concern, or exhibit an impolite tone. This annotated dataset reveals some recurring issues, and the team can adjust the model's training data to address these weaknesses.

\subsubsection{Manual editing}
To make the annotation even more substantial, users can edit the model's response to leave notes useful for further analysis. For instance, they can mark with some tokens a specific place where the model's response was impolite.

\subsection{Prompt adjustment}

\subsubsection{Setup}
In this section, we use the Inference page of the tool, MATH~\citep{math} dataset and prerelease version of OpenMath-Mistral-7B-v0.1 model~\citep{openmath}. The MATH dataset consists of 12,500 challenging competition mathematics problems.

\subsubsection{Interactive inference}

While evaluating the performance of the OpenMath-Mistral-7b model on the MATH dataset, we observed instances where the model failed to recall or apply certain common facts. This prompted us to hypothesize that reinforcing these facts might improve the model's performance on the dataset. To test this hypothesis, we implemented the following approach:

\begin{enumerate}
    
\item  We chose a single question from the dataset where the model could not solve the question due to the lack of knowledge, i.e.:

\begin{quote}
\textit{Let $a,$ $b,$ $c$ be the roots of $3x^3 - 3x^2 + 11x - 8 = 0.$  Find $ab + ac + bc.$}
\end{quote}

To solve the question, the model should know Vieta's formula.

\item We added information about the missing fact to the prompt:

\begin{quote}
You're an expert mathematician.
Help the user to solve this problem.

\textit{You can use the following facts to solve the problem:}

\textit{Vieta’s Formula for the Cubic Equation: If $f(x) = a_3x^3 + a_2x^2 + a_1x + a_0$ is a quadratic equation with roots $\alpha$, $\beta$ and $\gamma$ then,
sum of roots = $\alpha + \beta + \gamma = -\frac{a_2}{a_3}$;
sum of product of two roots = $\alpha\beta + \alpha\gamma + \beta\gamma = \frac{a_1}{a_3}$;
product of roots = $\alpha\beta\gamma = -\frac{a_0}{a_3}$.}

\end{quote}

\item Sent the updated query to the model and checked whether the model could solve the problem with our help - it could successfully apply the formula and solve the problem.

\end{enumerate}

The inference page can also be used for initial prompt adjustment. When using a new model for synthetic data generation, it is crucial to verify the effectiveness of the prompt and adjust it as needed to meet the generation goals.
\section{Results}
\label{section:results}
\subsection{GSM-Plus Cleaning}
We used the tool to reduce bad quality samples in the synthetic GSM-Plus dataset from 46.99\% to 19.51\% - less than 5\% in each category. To assess the quality of the initial and cleaned versions of the GSM-Plus dataset, we manually checked ten random samples per perturbation category where the model generated incorrect responses. The cleaned dataset can be found in the NeMo-Skills repository.~\footnote{\url{https://github.com/NVIDIA/NeMo-Skills/blob/main/nemo_skills/dataset/gsm-plus/prepare.py}}

In section~\ref{gsm_plus}, we show only a few examples of GSM-Plus dataset analysis. To get the final version we cleaned each category deeper.

\subsection{OpenMathInstruct-1}
We applied NeMo-Inspector to identify error patterns in OpenMath-Mistral-7B-v0.1 models on GSM8K and MATH~\citep{math} datasets. The error analysis revealed some recurrent problems: 1) \textit{Code Execution Errors}, when the models generate code that executes with an error message, the models cannot recover from it. The tool's statistics feature helped identify approximately 26\% of such execution-related issues across code-based solutions. 2) \textit{Calculation Errors}, when models perform correct reasoning steps but fail in arithmetic calculation. These errors were identified by comparing text-based solutions with their corresponding code-based solutions, which did not exhibit the same issue. There are in-depth examples of these errors in~\citep{openmath}. 
To address code execution errors, we relaunched the code generation when we encountered an error. This led to a 1.92\% accuracy improvement on OpenMath models on the MATH dataset. To rectify calculation issues, we updated the dataset used for fine-tuning. We changed all expressions where the model combines many arithmetic operations into one at a time and changed the single combined operation to many small ones. This boosted the accuracy of the Meta-Llama-3-8B model (fine-tuned on Nemotron-340B~\citep{nemotron} generated data) on the GSM8K dataset by 4.17\%. Our results underscore the importance of data exploration in the development of language models.

\section{Conclusion}
\label{section:conclusion}
This paper introduces NeMo-Inspector, an open source tool developed to facilitate the analysis of LLM generations and experimentation with various inference configurations. 
It offers simultaneous analysis from different perspectives by allowing to work with both \textit{homogeneous} and \textit{heterogeneous} LLM generations.
NeMo-Inspector features advanced visualization capabilities and provides streamlined inference, as well as user-defined custom functions and statistics.

NeMo-Inspector effectively expedited the cleaning process of the GSM-Plus dataset, reducing the number of low-quality samples from 46.99\% to 19.51\%. Furthermore, the tool helped detect generation issues in OpenMath models that led to a 1.92\% drop in accuracy for these models on the MATH dataset and 4.17\% on the GSM8K dataset for Meta-Llama-3-8B model fine-tuned on data generated from Nemotron-4-340b.~\citep{Ando2005}

\section{Limitations}
\label{section:limitations}
The current version has the following  limitations:
\begin{itemize}
\item The tool is designed to facilitate the analysis of any text datasets across various domains. However, its functionality is restricted to datasets formatted in the JSON Lines (JSONL) structure.

\item The tool lacks hosting capabilities and instead executes code written by users in custom functions. This approach poses potential security risks, particularly when the tool is made available for use by external users.

\item Inference works through NeMo-Skills scripts. Therefore, it supports only models supported by NeMo-Skills scripts.
\end{itemize}

\section*{Acknowledgments}
The authors would like to thank Elena Rastorgueva, Grigor Nalbandyan and Ivan Moshkov for their review and feedback.

\bibliography{main}

\end{document}